\documentclass[10pt,twocolumn,letterpaper]{article}

\usepackage{iccv}
\usepackage{times}
\usepackage{epsfig}
\usepackage{graphicx}
\usepackage{amsmath}
\usepackage{amssymb}
\usepackage{bm}
\usepackage{booktabs}
\usepackage[ruled,lined,boxed,linesnumbered]{algorithm2e}
\usepackage[american]{babel}
\usepackage{multirow} 
\usepackage{subfig}

\usepackage[pagebackref=true,breaklinks=true,letterpaper=true,colorlinks,bookmarks=false]{hyperref}

\iccvfinalcopy 

\def\iccvPaperID{2263} 

\ificcvfinal\fi
\begin{document}

\title{Image Harmonization Dataset iHarmony4: HCOCO, HAdobe5k, HFlickr, and Hday2night}
\author{$\textnormal{Wenyan Cong}^{*}$, $\textnormal{Jianfu Zhang}^{*}$, $\textnormal{Li Niu}^{*}$, $\textnormal{Liu Liu}^{*}$, $\textnormal{Zhixin Ling}^{*}$, $\textnormal{Weiyuan Li}^{\dagger}$, $\textnormal{Liqing Zhang}^{*}$\\
$^*$ Shanghai Jiao Tong University 
$^\dagger$ East China Normal University
}

\maketitle
\thispagestyle{empty}

\begin{abstract}
Image composition is an important operation in image processing, but the inconsistency between foreground and background significantly degrades the quality of composite image. Image harmonization, which aims to make the foreground compatible with the background, is a promising yet challenging task. However, the lack of high-quality public dataset for image harmonization, which significantly hinders the development of image harmonization techniques. Therefore, we contribute an image harmonization dataset iHarmony4 by generating synthesized composite images based on existing COCO (\emph{resp.}, Adobe5k, day2night) dataset, leading to our HCOCO (\emph{resp.}, HAdobe5k, Hday2night) sub-dataset.  To enrich the diversity of our dataset, we also generate synthesized composite images based on our collected Flick images, leading to our HFlickr sub-dataset. The image harmonization dataset iHarmony4 is released in \href{https://github.com/bcmi/Image_Harmonization_Datasets}{https://github.com/bcmi/Image\_Harmonization\_Datasets}.
\end{abstract}

\begin{figure*}[t]
\subfloat[Microsoft COCO $\&$ Flickr]{
\label{Fig:R1}
\begin{minipage}{0.5\textwidth}
\centering
\includegraphics[width=8cm]{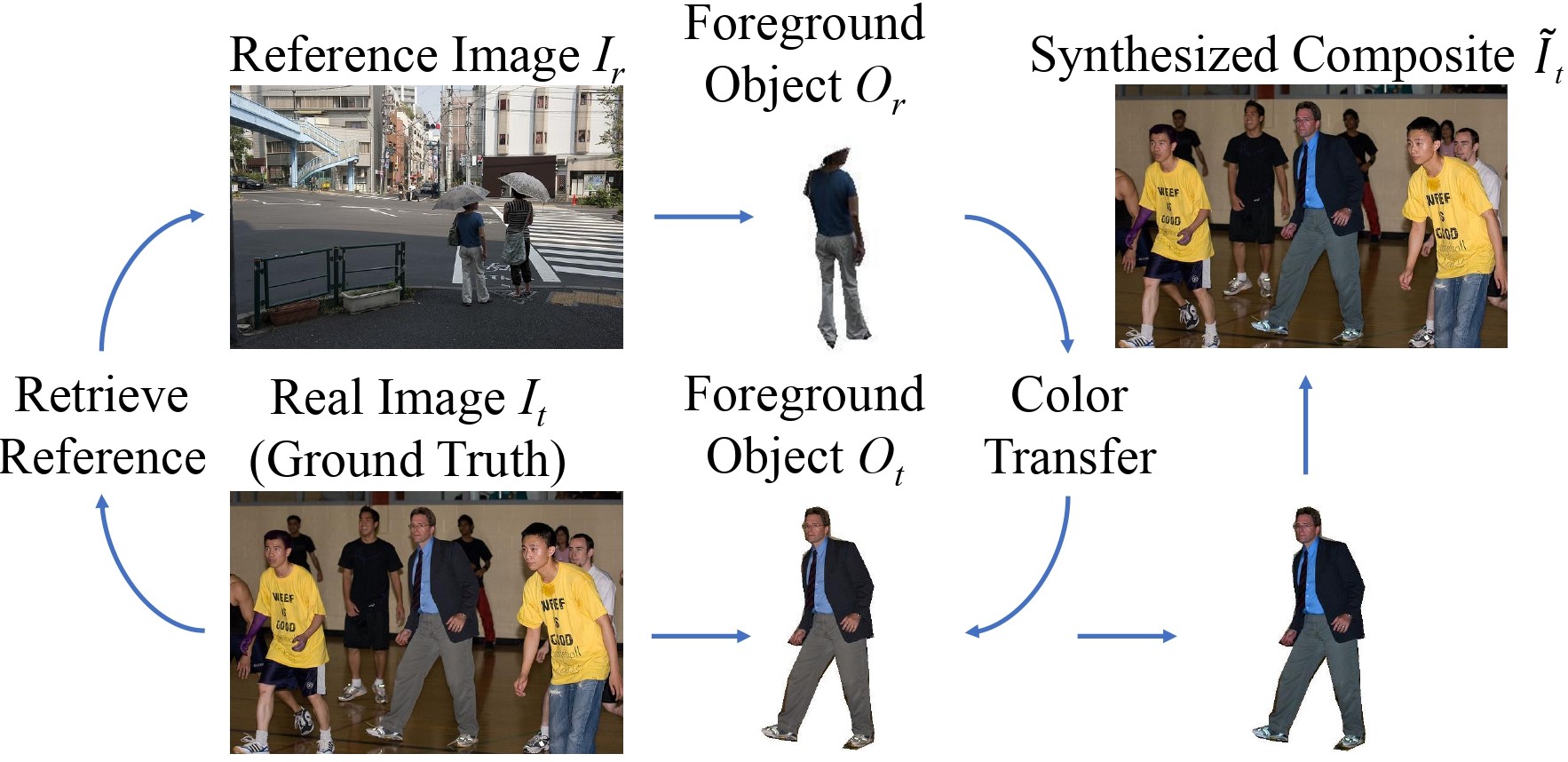}
\end{minipage}
}
\subfloat[MIT-Adobe Fivek $\&$ day2night]{
\label{Fig:R2}
\begin{minipage}{0.5\textwidth}
\centering
\includegraphics[width=8cm]{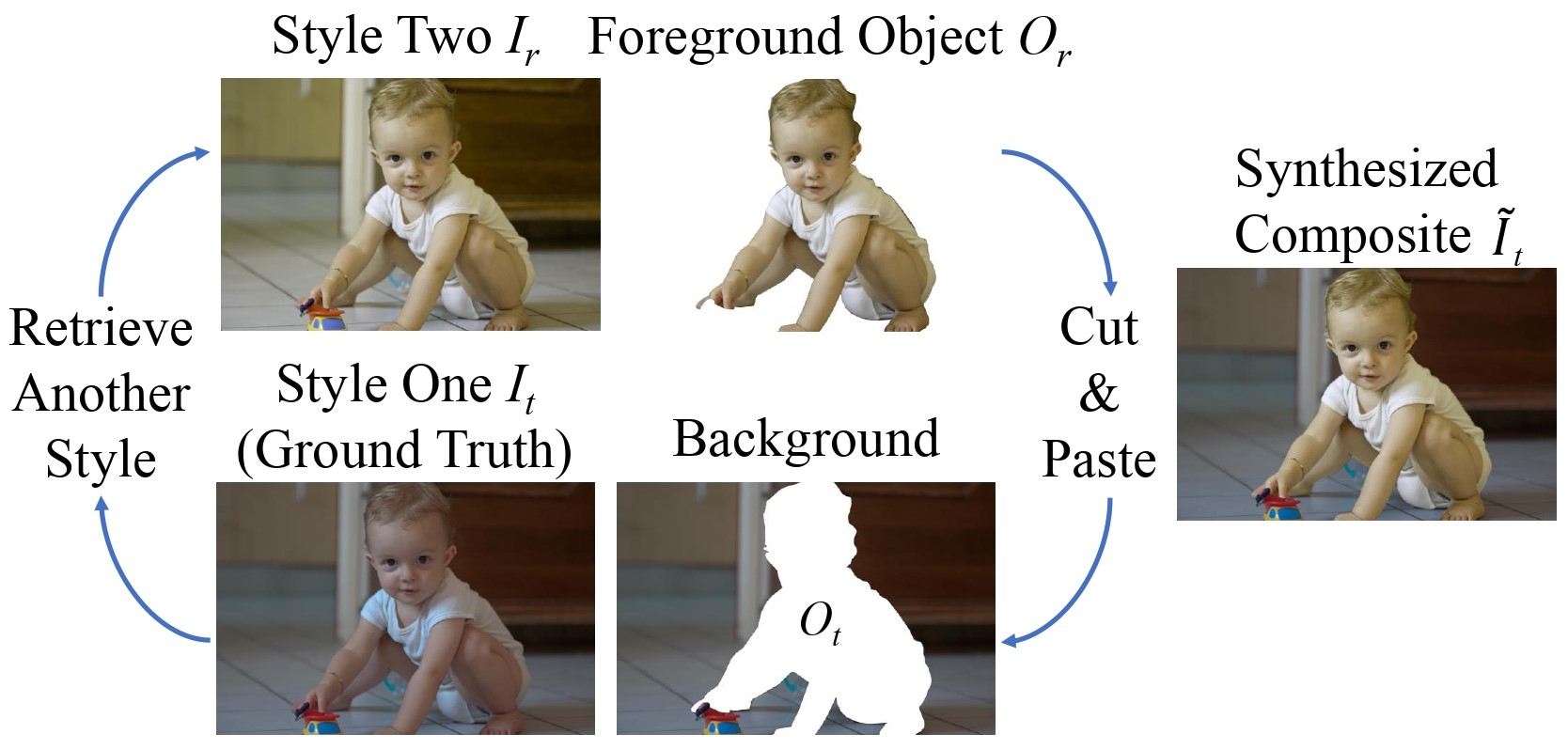}
\end{minipage}
}
  \caption[]{The illustration of data acquisition process. \subref{Fig:R1} On Miscrosoft COCO and Flickr datasets, given a target image $I_t$ with foreground object $O_t$, we find a reference image $I_r$ with foreground object $O_r$ from the same category as $O_t$, and then transfer color information from $O_r$ to $O_t$. \subref{Fig:R2} On MIT-Adobe5k and day2night datasets, given a target image $I_t$ with foreground object $O_t$, we find its another version $I_r$ (edited to form a different style or captured in a different illumination condition) and overlay $O_t$ with the corresponding $O_r$ at the same location in $I_r$.}
  \label{fig:process}
\end{figure*}

\section{Introduction} \label{sec:intro}

Image composition aims to generate a composite image by extracting the foreground of one image and paste it on the background of another image. However, since the foreground is usually not compatible with the background, the quality of composite image is significantly downgraded. To address this issue, image harmonization aims to adjust the foreground to make it compatible with the background in the composite image.  Both traditional methods~\cite{lalonde2007using,xue2012understanding,zhu2015learning} and deep learning based method~\cite{tsai2017deep} have been explored for image harmonization, in which deep learning based method~\cite{tsai2017deep} could achieve promising results. 

As a data-hungry approach, deep learning requires a large number of training pairs of composite image and harmonized image as input image and its ground-truth output. However, given a composite image, manually creating its harmonized image, \ie, adjusting the foreground to be compatible with background, is in high demand for extensive efforts of skilled expertise. So this strategy of constructing datasets is very time-consuming and expensive, making it infeasible to generate large-scale training data. Alternatively, as proposed in~\cite{tsai2017deep}, we can treat a real image as harmonized image, segment a foreground region, and adjust this foreground region to be inconsistent with the background, yielding a synthesized composite image. Then, pairs of synthesized composite image and real image can be used to supersede pairs of composite image and harmonized image. Since foreground adjustment can be done automatically (\emph{e.g.}, color transfer methods), time-consuming expertise editing is not required, which makes it feasible to collect large-scale training data. Although the work in~\cite{tsai2017deep} proposed an inspiring strategy, it does not make its constructed datasets available.

Therefore, we adopt the strategy in~\cite{tsai2017deep} to generate pairs of synthesized composite image and real image. Moreover, we tend to release our constructed dataset iHarmony4 to facilitate the research in the field of image harmonization. Similar to~\cite{tsai2017deep}, we generate synthesized composite images based on Microsoft COCO dataset~\cite{lin2014microsoft}, MIT-Adobe5k dataset~\cite{bychkovsky2011learning}, and self-collected Flickr dataset. For Flickr dataset, we crawl images from Flickr image website by using the list of category names in ImageNet dataset~\cite{imagenet_cvpr09} as queries, in order to increase the diversity of crawled images. However, not all crawled images are suitable for the image harmonization task.  So we manually filter out the images with pure-color or blurry background, the cluttered images with no obvious foreground objects, and the images which appear apparently unrealistic due to artistic editing.

Besides COCO, Adobe5k, and Flickr suggested in~\cite{tsai2017deep}, we additionally consider datasets which capture multiple images in different illumination conditions for the same object or scene. Such datasets are naturally beneficial for image harmonization task because composite images can be easily generated by replacing the foreground region in one image with the same foreground region in another image. More importantly, two foreground regions are both from real images, and thus the composite image is real composite image. However, to the best of our knowledge, there are only a few available datasets~\cite{shih2013data, zhou2016evaluating,Laffont14} in this scope. 
Finally, we choose day2night dataset~\cite{Laffont14}, because day2night provides a collection of aligned images captured in a variety of environments (\emph{e.g.}, weather, season, time of day) for each scene.

According to the names of original datasets, we refer to the constructed sub-datasets as HCOCO, HAdobe5k, HFlickr, and Hday2night separately, in which H stands for harmonization. And the whole dataset is referred to as iHarmony4. The details of constructing dataset iHarmony4 and the difference from \cite{tsai2017deep} will be described in the following sections.

\begin{figure*}[t]
\centering
  \includegraphics[width=17cm]{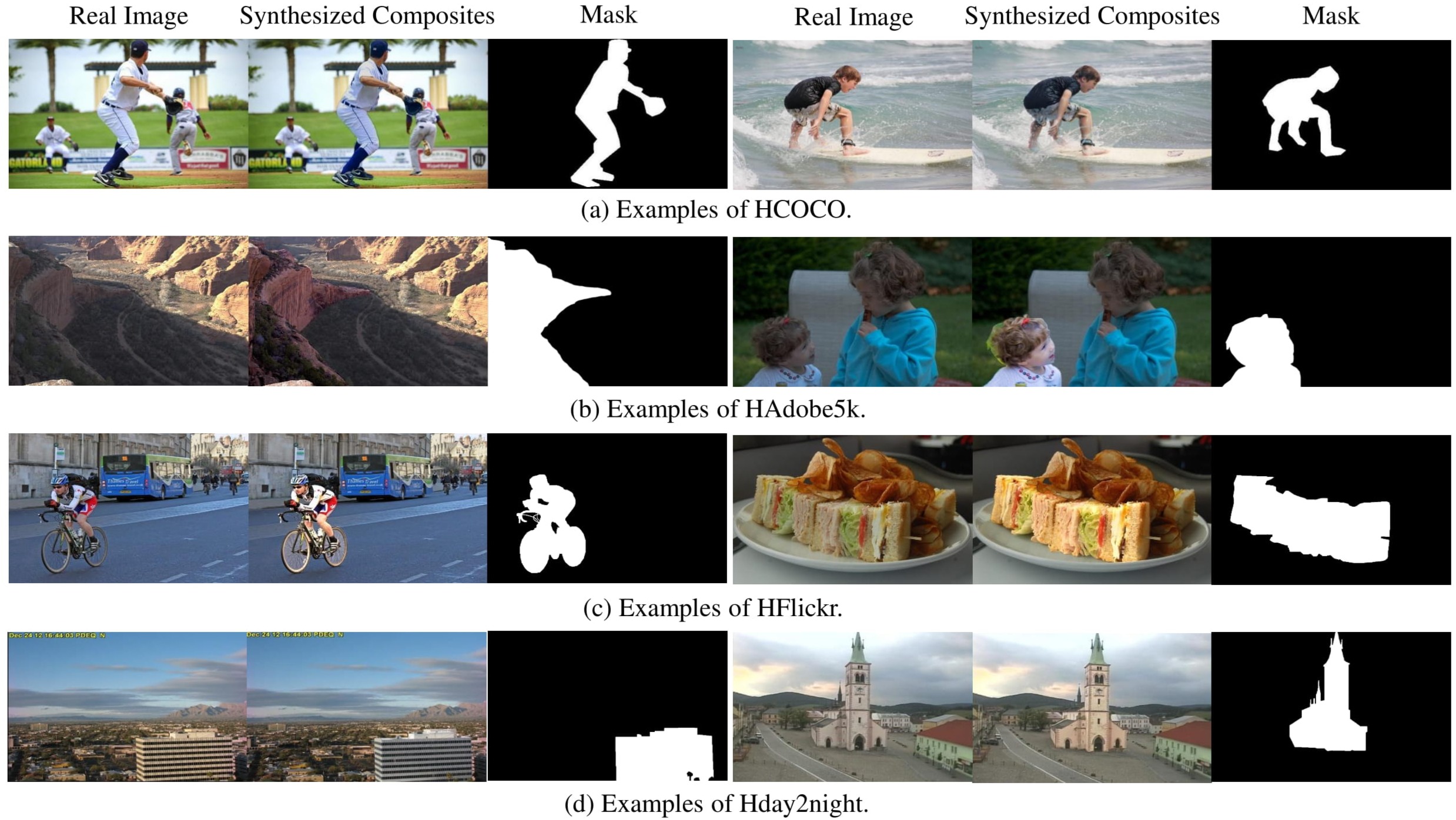}
  \caption[]{Example images of our contributed dataset iHarmony4. From top to bottom, we show examples from our HCOCO, HAdobe5k, HFlickr, and Hday2night sub-datasets. From left to right, we show the real image, the synthesized composite image, and the foreground mask for each example.}
  \label{fig:samples}
\end{figure*}

\section{Synthesized Composite Image Generation}

The process of generating synthesized composite image from a real image can be divided into two steps: foreground segmentation and foreground adjustment.

\subsection{Foreground Segmentation}
For COCO dataset, we use the provided segmentation mask.  The other datasets (\emph{i.e.}, Adobe5k, Flickr, and day2night) are not associated with segmentation masks, so we manually segment one or more foreground regions for each image. We do not use pretrained segmentation model to automatically segment foreground regions to ensure the accuracy of obtained segmentation masks. 

On all four sub-datasets,  we ensure that each foreground region occupies a reasonable area of the whole image. When selecting the foreground region, we also attempt to make the foreground objects cover a wide range of categories.

\subsection{Foreground Adjustment}

After segmenting a foreground region $O_t$ in one image $I_t$, we need to adjust the appearance of $O_t$. For ease of description, $I_t$ is referred to as target image. As suggested in \cite{tsai2017deep}, another image $I_r$ containing the foreground region $O_r$ is chosen as reference image. Then, color information is transferred from $O_r$ to $O_t$, leading to a synthesized composite image $I_t'$. This process is illustrated in Figure ~\ref{fig:process}. 

For Adobe5k dataset, each real image is retouched by five professional photographers using Adobe Lightroom, so one real target image $I_t$ is accompanied by five edited images $\{I_i|_{i=1}^5\}$ in different styles. We could randomly select $I_r$ from $\{I_i|_{i=1}^5\}$ and overlay $O_t$ in $I_t$ with the corresponding region $O_r$ at the same location in $I_r$.

For day2night dataset, each scene is captured in different environments, resulting in a set of aligned images $\{I_i|_{i=1}^n\}$. Similar to Adobe5k, a target image $I_t$ and a reference image $I_r$  could be randomly selected from $\{I_i|_{i=1}^n\}$, followed by overlaying $O_t$ in $I_t$ with the corresponding region $O_r$ at the same location in $I_r$. However, different from Adobe5k, we need to make sure that $O_t$ and $O_r$ are the same object without essential change. For example, moving objects (\emph{e.g.}, person, animal, car) in $I_t$ may move or disappear in  $I_r$. Besides, even the static objects (\emph{e.g.} building, mountain) in $I_t$ may be different from those in $I_r$, like building with lights on in $I_t$ while lights off in $I_r$. The above foreground changes come from the objects themselves instead of outside environment, and thus those pairs are eliminated from our dataset.

For COCO and Flickr datasets, since they do not have aligned images, given a target image $I_t$ with foreground $O_t$, we randomly select a reference image $I_r$ with foreground $O_r$ belonging to the same category as $O_t$. More specifically, as COCO dataset provides segmentation masks and category information, we naturally leverage its annotation to search for proper reference. For Flickr, since there is no semantic information provided, we first use a scene parsing model~\cite{sceneparsing} pretrained on ADE20K~\cite{ade20k} to predict pixel-wise segmentation labels of each image. Given a target foreground, we obtain its dominant category labels based on predicted pixel-wise labels, and retrieve references containing the objects of the same category in ADE20K dataset. Then, as suggested in \cite{tsai2017deep}, we apply color transfer method to transfer color information from $O_r$ to $O_t$. Nevertheless, the work \cite{tsai2017deep} only utilizes one color transfer method~\cite{lee2016automatic}, which limits the diversity of generated images. Considering that color transfer methods can be categorized into four groups based on parametric/non-parametric and correlated/decorrelated color space, we select one representative method from each group, \emph{i.e.},  parametric method~\cite{reinhard2001color} in decorrelated color space, parametric method~\cite{xiao2006color} in correlated color space, non-parametric~\cite{fecker2008histogram} in decorrelated color space, and non-parametric~\cite{pitie2007automated} in correlated color space. Given a pair of $O_t$ and $O_r$, we randomly choose one from the above four color transfer methods.

\section{Synthesized Composite Image Filtering}
Through foreground segmentation and adjustment, we can obtain a large amount of synthesized composite images. However, some of the synthesized foreground objects look unrealistic, so we use aesthetics prediction model~\cite{kong2016photo} to remove unrealistic composite images. To further remove unrealistic composite images, we train a binary CNN classifier by using the real images as positive samples and the unrealistic composite images identified by~\cite{kong2016photo} as negative samples. When training the classifier, we also feed foreground mask into CNN to provide the foreground information. 

After two steps of automatic filtering, there are still some remaining unrealistic images. Thus, we ask human annotators to filter out the remaining unrealistic images manually. During manual filtering, we also consider another two critical issues: 1) for COCO dataset, some selected foreground regions are not very reasonable such as highly occluded objects, so we remove these images manually; 2) for COCO and Flickr datasets, the hue of some foreground objects are vastly changed after color transfer, which generally happens to the categories with large intra-class variance. For example, a red car is transformed into a blue car, or a man in red T-shirt is transformed into a man in green T-shirt. This type of color transfer is not very meaningful for image harmonization task, so we also remove these images manually.

\section{Dataset Statistics}
In this section, we introduce the details of our constructed dataset iHarmony4.

\noindent\textbf{HCOCO: }Microsoft COCO dataset ~\cite{lin2014microsoft}, containing 118k images for training and 41k for testing, is a large-scale dataset for object detection, segmentation, and captioning. It provides the object segmentation masks for each image with 80 object categories annotated in total. To generate more convincing composites, training set and test set are merged together to guarantee a wider range of available references. Based on COCO dataset, we build our HCOCO sub-dataset with 42828 pairs of synthesized composite image and real image.

\noindent\textbf{HAdobe5k: }MIT-Adobe5k dataset~ \cite{bychkovsky2011learning} covers a wide range of scenes, objects, and lighting conditions. For all the 5000 photos, each of them is retouched by five photographers, producing five different renditions, which are supposed to be visually pleasing and realistic. We use 4329 images with one segmented foreground object in each image to build our HAdobe5k sub-dataset, which has 21597 pairs of synthesized composite image and real image.

\noindent\textbf{HFlickr: }Flickr contains diverse images uploaded by amateur photographers. With searchable metadata and keywords, it is feasible to crawl a diversity of images using category names as queries. We construct our HFlickr sub-dataset based on crawled 4833 Flickr images with one or two segmented foreground objects in each image, and obtain 8277 pairs of synthesized composite image and real image.

\noindent\textbf{Hday2night: }Day2night dataset~\cite{zhou2016evaluating} collected from AMOS dataset~\cite{AMOSdataset} contains images taken at different times of the day by fixed webcams. There are 8571 images of 101 different scenes in total. We select 106 target images from 80 scenes with one segmented foreground object in each image to generate composites. Our Hday2night sub-dataset has 444 pairs of synthesized composite image and real image.

\begin{table}
\centering
\caption{Number of training and test images of four synthesized sub-datasets in iHarmony4.}
\resizebox{0.9\columnwidth}!{
\begin{tabular}{ccccc}\\
\toprule
 Sub-dataset & HCOCO & HAobe5k & HFlickr & Hday2night\\
\midrule
\#Training set & 38545 & 19437 & 7449 & 311 \\
\#Test set & 4283 & 2160 & 828 & 133 \\
\bottomrule
\end{tabular}}
\label{statistics}
\end{table}

For each constructed sub-dataset (\ie, HCOCO, HAobe5k, HFlickr, and Hday2night), all pairs are split into training set and test set. We ensure that the same target image does not appear in the training set and test set simultaneously, to avoid that the trained model simply memorize the target image. The numbers of training and test images in four sub-datasets are summarized in Table~\ref{statistics}. In Figure~\ref{fig:samples}, we also show some sample images with real image, synthesized composite image, and foreground mask from each sub-dataset in iHarmony4.

\section{Conclusions}
We have constructed an image harmonization dataset iHarmony4 with four sub-datasets, \emph{i.e.}, HCOCO, HAobe5k, HFlickr, and Hday2night, which cover a wide range of object categories and color transfer methods. We have also used automatic filtering and manual filtering to guarantee the high quality of synthesized composite images.

\bibliographystyle{ieee}
\bibliography{egbib}

\end{document}